\documentclass[conference]{IEEEtran}
\IEEEoverridecommandlockouts
\usepackage{cite}
\usepackage{amsmath,amssymb,amsfonts}
\usepackage{algorithmic}
\usepackage{graphicx}
\usepackage{textcomp}
\usepackage{booktabs}
\usepackage{multirow}
\usepackage{multicol}
\usepackage{xcolor}
\usepackage{url}
\def\BibTeX{{\rm B\kern-.05em{\sc i\kern-.025em b}\kern-.08em
    T\kern-.1667em\lower.7ex\hbox{E}\kern-.125emX}}
\begin{document}

\title{SRL-MAD: Structured Residual Latents for One-Class Morphing Attack Detection}

\author{
Diogo J. Paulo$^{1,2,3}$, Hugo Proença$^{1,2}$, João C. Neves$^{1,3}$\\[4pt]
$^{1}$University of Beira Interior, Portugal \quad $^{2}$IT: Instituto de Telecomunicações \quad $^{3}$NOVA LINCS\\
diogo.paulo@ubi.pt
}

\maketitle

\begin{abstract}
Face morphing attacks represent a significant threat to biometric systems as they allow multiple identities to be
combined into a single face. While supervised morphing attack detection (MAD) methods have shown promising performance, their reliance on attack-labeled data limits generalization to unseen morphing attacks. This has motivated increasing interest in one-class MAD, where models are trained exclusively on bona fide samples and are expected to detect unseen attacks as deviations from the normal facial structure. In this context, we introduce SRL-MAD, a one-class single-image MAD that uses structured residual Fourier representations for open-set morphing attack detection. Starting from a residual frequency map that suppresses image-specific spectral trends, we preserve the two-dimensional organization of the Fourier domain through a ring-based representation and replace azimuthal averaging with a learnable ring-wise spectral projection. To further encode domain knowledge about where morphing artifacts arise, we impose a frequency-informed inductive bias by organizing spectral evidence into low, mid, and high-frequency bands and learning cross-band interactions. These structured spectral features are mapped into a latent space designed for direct scoring, avoiding the reliance on reconstruction errors. Extensive evaluation on FERET-Morph, FRLL-Morph, and MorDIFF demonstrates that SRL-MAD consistently outperforms recent one-class and supervised MAD models. Overall, our results show that learning frequency-aware projections provides a more discriminative alternative to azimuthal spectral summarization for one-class morphing attack detection.
\end{abstract}


\section{Introduction}
Face morphing attacks represent a significant threat to biometric authentication systems since they allow adversaries to bypass face recognition by blending facial features from multiple individuals. As a result, single-image morphing attack detection (S-MAD) has been investigated to enhance the security of biometric pipelines. However, despite progress in supervised S-MAD, real-world deployment remains difficult due to the diversity of morphing techniques and the frequent occurrence of previously unseen attack distributions as novel morphing methods are developed. This gap has motivated the interest in one-class and anomaly-based S-MAD, in which models are trained exclusively on bona fide samples and are expected to detect previously unseen morphing attacks as deviations from normal patterns~\cite{mad_ddpm, spl-mad, colbois2024evaluating}. These methods reduce the dependence on specific attack models and are better aligned with open-set and cross-dataset evaluation protocols. Nonetheless, the development of such methods that are both interpretable and robust remains a challenge, particularly when dealing with high-dimensional facial representations. Moreover, existing one-class models are sensitive to the quality and size of the training data~\cite{mad_ddpm}, struggling to learn representations that explicitly separate bona fide images from morph-specific artifacts~\cite{damer2019generalization}.

\begin{figure}[t]
    \vspace{0.5em}
    \centering
    \includegraphics[width=\linewidth]{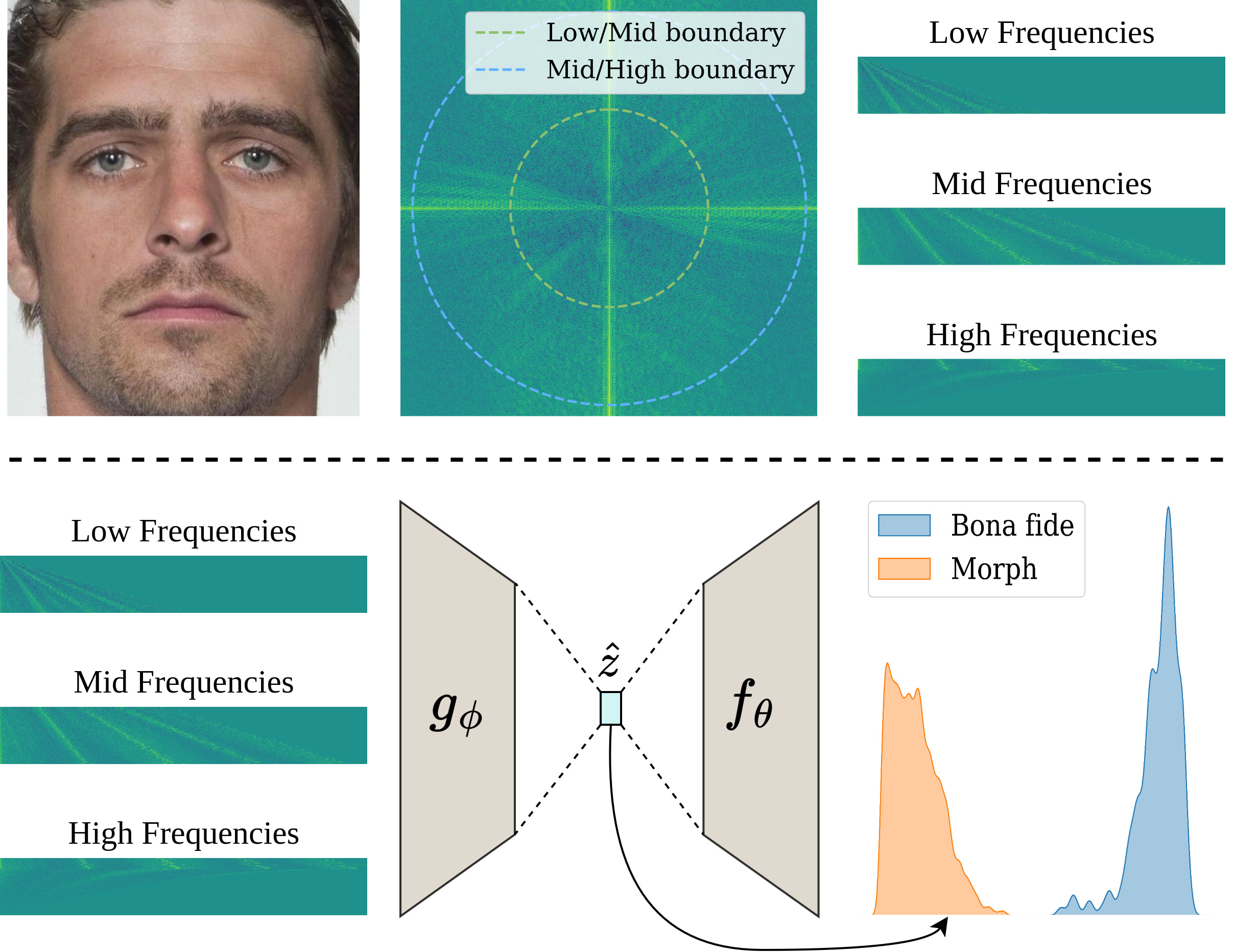}
    \caption{\textbf{Illustration of SRL-MAD.} From a single face image, residual Fourier features are extracted and divided into low, mid, and high-frequency bands. These are projected into a compact 1D latent space where morphing attacks appear well-separated from bona fides.}
    \label{fig:teaser_placeholder}
\end{figure}

Motivated by the current challenges, we propose SRL-MAD, a one-class method that transforms structured residual Fourier features, extracted from facial images, into an interpretable discriminative score. Rather than relying on reconstruction errors, which have been shown to be unstable and attack-dependent~\cite{spl-mad}, the proposed method learns a frequency-structured latent representation that captures bona fide residual patterns. Therefore, morphing attacks are detected as deviations from this learned manifold. Our approach starts by obtaining a residual Fourier map, from a single face image, by removing the image-specific power-law spectrum $f^{-\alpha}$, emphasizing deviations associated with morphing artifacts~\cite{paulo2026fdmadfrequencydomainresidualanalysis}. Rather than collapsing this residual map using the azimuthal average technique (i.e., radially averaging frequency magnitudes and discarding angular structure), we preserve its two-dimensional organization by dividing it into concentric rings and introducing a learnable ring-wise spectral projection mechanism, which enables adaptive weighting of frequency components within each spatial ring and eases more expressive modeling of residual statistics while preserving the geometric structure. Furthermore, we introduce a structured frequency prior by grouping rings into low, mid, and high-frequency bands, which are projected into a compact representation that captures cross-band interactions. Projecting all bands into these three bands allows the model to jointly reason about global coherence (low bands) and localized morphing traces that typically manifest in the mid/high bands. In contrast to the azimuthal average, these bands are learned, creating a latent space that is enforced by mapping this representation to a single latent coordinate, serving as a discriminative score where morphing attacks appear well-separated from bona fides (Figure~\ref{fig:teaser_placeholder}). Anomaly scoring is performed in the latent space rather than the reconstruction space, thereby improving stability across morphing techniques and datasets. We evaluate the proposed method on three common benchmarks, including FERET-Morph, FRLL-Morph~\cite{frll_morph}, and MorDIFF~\cite{mordiff}, which cover landmark-based, GAN-based, and diffusion-based morphing attacks.

Our method achieves an average Equal Error Rate (EER) of 4.76\% across all test sets, outperforming recent one-class approaches such as SPL-MAD~\cite{spl-mad} and MAD-DDPM~\cite{mad_ddpm}, and remains competitive with SelfMAD++~\cite{selfmad++}. Additionally, our approach surpasses several supervised methods, including MADation~\cite{caldeira2025madation}, achieving a lower average EER. These results show that using structured frequency-domain residuals and enforcing a compact latent space provide a strong foundation for open-set morphing attack detection, advancing the state of the art in one-class S-MAD.

In summary, our  contributions are three-fold:
\begin{itemize}
    \item \textbf{Frequency-aware modeling for one-class morphing attack detection.} We introduce SRL-MAD that transforms residual Fourier features into a compact latent manifold explicitly separating bona fide images from morphs.

    \item \textbf{Learnable geometry-preserving spectral aggregation.} We replace azimuthal averaging with learnable ring-wise spectral projection, enabling adaptive weighting of frequency components while preserving the two-dimensional organization of the Fourier magnitude map.

    \item \textbf{Frequency-domain inductive bias with cross-band interaction modeling.} We encode prior knowledge on where morphing artifacts manifest by organizing residual spectra into low, mid, and high-frequency groups and learn a joint projection that captures dependencies between global structure (low) and localized blending/texture inconsistencies (mid/high).

\end{itemize}

\section{Related Work}
\vspace{0.5em}
\noindent\textbf{Morphing Attack Detection.} Early approaches of face morphing attack detection used handcrafted features such as LBP, LPQ, BSIF, or SIFT descriptors to capture subtle morphing artifacts~\cite{texture_morphs, texture_morphs2}, but showed limited generalization to unseen morphing techniques. In contrast, deep learning-based S-MAD~\cite{orthomad, pwmad, idistill, caldeira2025madation, pca_morph, transformermad} approaches have achieved higher performance since they learned hierarchical feature representations. Recently, foundation model-based approaches such as MADation~\cite{caldeira2025madation} adapted CLIP-like vision-language architectures~\cite{clip} for MAD tasks, exploiting large-scale pretraining to enhance zero-shot generalization to unseen morphing techniques. However, most of these supervised MAD methods still rely on labeled morphing datasets and exhibit limited cross-domain robustness. To address generalization gaps, SelfMAD~\cite{selfmad} introduced a self-supervised approach to simulate morphing artifacts in both pixel and frequency domains. Its successor, SelfMAD++~\cite{selfmad++}, further integrates foundation models with local feature enhancement and self-supervised pretraining, improving generalization to unseen morphing types through localized artifact reasoning and representation learning. Additionally, recent work has explored alternative input representations that use residual images or frequency-domain representations~\cite{ramachandra2019detecting, venkatesh2019morphed, pca_morph, huber2025fx, spl-mad} instead of spatial-domain RGB images. These methods aim to emphasize subtle artifacts introduced by morphing while suppressing identity-related information. Our work follows this line by operating on structured residual spectra, but differs in its focus on anomaly detection rather than supervised classification.

\begin{figure*}[ht]
    \centering
    \includegraphics[width=\linewidth]{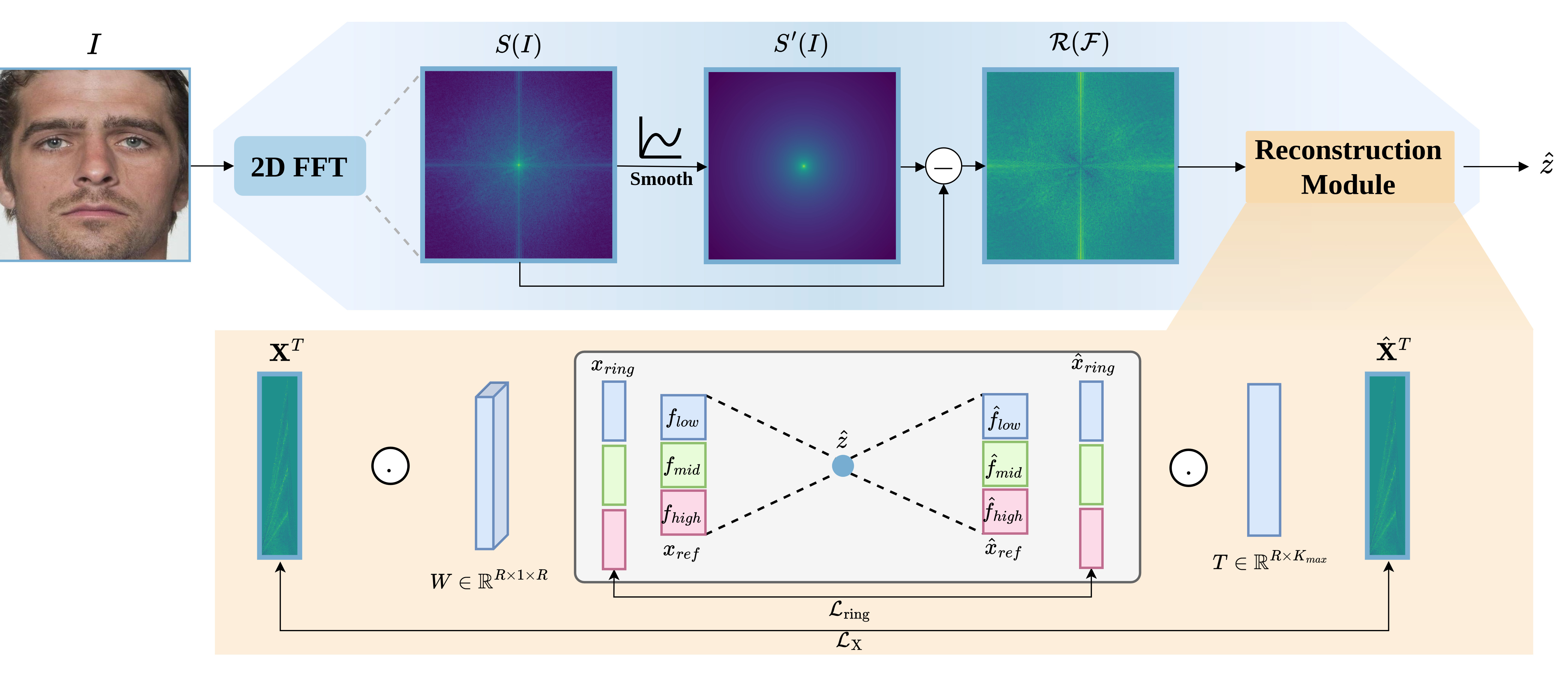}
    \caption{\textbf{Overview of the proposed method (SRL-MAD).} We extract the Residual Fourier Map from a single face image by subtracting a power-law baseline to highlight morphing artifacts. The map is divided into concentric rings to form a structured 2D feature representation, which is then processed through a learnable ring-wise spectral projection mechanism. These features are refined into low, mid, and high-frequency bands and projected by an autoencoder into a compact 1D latent space. The final 1D latent coordinate serves as a discriminative score.}
    \label{fig:method_placeholder}
\end{figure*}
\vspace{0.5em}
\noindent\textbf{Anomaly-Based and One-Class MAD.} Anomaly detection has emerged as a promising direction for MAD, modeling the distribution of bona fide samples and treating morphs as outliers. In~\cite{spl-mad}, the authors introduced self-paced learning to separate bona fide and morphed samples through unsupervised training, improving generalization and reconstruction separability, and showed that morphing attacks can be easier to reconstruct than bona fide images, as evidenced by lower reconstruction errors. Furthermore,~\cite{mad_ddpm} proposed MAD-DDPM, a diffusion-based one-class model that learns bona fide image distributions using denoising diffusion probabilistic models~\cite{ddpm}. Despite these advances, these approaches typically rely on reconstruction errors as anomaly scores. However, reconstruction errors are often sensitive to architectural choices and may fail when models generalize well to both bona fide and attack samples~\cite{spl-mad}. Additionally, \cite{colbois2024evaluating} explored attack-agnostic features extracted by large vision models pretrained only on bona fide data. These features, modeled with a Gaussian mixture model, outperform traditional one-class detectors across unseen morph scenarios. Our method aligns with this line of work but explicitly enforces a compact latent space in a one-dimensional space, making deviations from the bona fide distribution directly interpretable and robust. The anomaly score is derived from the bona fide latent distribution, avoiding reliance on reconstruction error at test time.

\vspace{0.5em}
\noindent\textbf{Frequency-Domain cues in MAD.} Frequency-domain analysis has proven effective in face morphing detection, as morphing operations often introduce characteristic artifacts in specific frequency bands~\cite{huber2025fx, tapia_fourier}. Prior works have employed Fourier analysis, wavelet decompositions, and radial or azimuthal averages to summarize frequency information~\cite{wb_avcivas, paulo2026fdmadfrequencydomainresidualanalysis}. While the azimuthal average provides a compact representation, it assumes uniform importance across frequency components and spatial regions. Building on these insights, we improve frequency-based MAD by replacing azimuthal average with learnable ring-wise spectral projections, allowing the network to weight frequency components differently for each spatial ring. Furthermore, by combining rings into low, mid, and high-frequency groups, the model captures structured spectral dependencies while maintaining low dimensionality.

\section{Methodology}
\label{sec:method}
Figure~\ref{fig:method_placeholder} presents an overview of the proposed method. Our approach operates in the frequency domain and is based on residual Fourier analysis to extract features, which are subsequently modeled using an autoencoder. While inspired by prior work on global Fourier residual representations~\cite{paulo2026fdmadfrequencydomainresidualanalysis}, SRL-MAD goes beyond feature extraction, preserving the two-dimensional frequency structure through a learnable ring-wise spectral projection that replaces the azimuthal average and introduces frequency priors to guide the latent representation used for the score.

\subsection{Residual Fourier Map}
\label{subsec:residual}

Given an input facial image $I$, we obtain the two-dimensional discrete Fourier transform and keep the magnitude spectrum. To reduce dynamic range, a logarithmic compression is applied, obtaining:
\begin{equation}
S(u,v) = \log\!\left(1 + \left| \mathcal{F}\{I(x,y)\} \right| \right),
\end{equation}
where $\mathcal{F}(\cdot)$ denotes the 2D DFT mapping the spatial signal $I(x,y)$ to frequency coordinates $(u,v)$. Natural images approximately follow a power-law decay in the frequency domain. To suppress this dominant trend and highlight deviations associated with morph artifacts, we estimate a power-law baseline from the image itself. First, the frequency plane is divided into concentric radial bands, and a one-dimensional radial profile is obtained by averaging $S(u,v)$ within each band. A linear model is then fitted in log–log space:
\begin{equation}
\log S(f) \approx a + b \log f,
\end{equation}
where $f$ denotes the radial frequency associated with each band. This fitted baseline is subsequently mapped back to the two-dimensional frequency plane, producing a baseline map $S'$ that depends only on radial distance.

The final residual Fourier map $\mathcal{R}$ is obtained as the difference between the log-magnitude spectrum and the mapped baseline. The DC component is set to zero. It is worth mentioning that this residual is obtained per-image and does not involve subtracting statistics estimated from the training set.

\subsection{Ring-Based 2D Feature Representation}
\label{subsec:rings}
The residual Fourier map is divided into $R$ concentric rings centered at the DC component. Each ring collects frequency coefficients with a similar radial distance. Unlike earlier approaches that collapse each ring into a single value through azimuthal averaging~\cite{paulo2026fdmadfrequencydomainresidualanalysis}, i.e., averaging magnitudes over all angular directions at a fixed radius to produce a 1D radial profile, we keep all residual coefficients within each ring.

Since the number of frequency samples increases with radius, the number of coefficients per ring also varies. Let $K_r$ denote the number of coefficients in ring $r$, and define $K_{\max} = \max_r K_r$. We construct a fixed-size matrix $\mathbf{X} \in \mathbb{R}^{R \times K_{\max}}$, where each row corresponds to one ring and contains the residual coefficients ordered clockwise. Rings with fewer than $K_{\max}$ coefficients are padded with zeros and a corresponding binary mask $\mathbf{M}$ is saved to identify valid coefficients and ensure that padded positions do not influence learning. This representation preserves fine-grained frequency information within each band, avoiding azimuthal averaging and preserving intra-band structure that can be informative for morphing artifact detection.

\subsection{Learnable Ring Projection}
\label{subsec:projection}
The ring matrix $\mathbf{X}$ is first processed by a learnable ring-wise projection that combines frequency coefficients within each ring. Specifically, we learn a set of projection kernels that operate along the angular dimension, allowing the model to emphasize artifact-relevant spectral patterns while suppressing nuisance variability. This operation is performed as a depth-wise one-dimensional convolution with kernel size $K_{\max}$, yielding a compact ring descriptor $\mathbf{x}_{\text{ring}} \in \mathbb{R}^{R}.$ To prevent padded coefficients from affecting the learned projection weights, the binary mask is enforced throughout training, ensuring that gradients do not propagate through invalid positions. Compared to azimuthal averaging or linear dimensionality reduction (e.g., PCA), this formulation allows the model to adaptively emphasize discriminative frequency patterns.

\subsection{Frequency-Band Refinement}
\label{subsec:bands}
To encode a frequency-domain inductive bias informed by how morphing artifacts manifest across bands, the ring descriptor $x_{\text{ring}}$ is divided into three contiguous frequency bands corresponding to low, mid, and high frequencies, which are defined by equally dividing the radial index range. For each group, we learn a band-wise linear projection that combines the ring responses within the band. This results in a three-dimensional refined representation:
\begin{equation}
x_{\text{ref}} =
\big[
f_{\text{low}}(x_{\text{ring}}^{(1)}),
f_{\text{mid}}(x_{\text{ring}}^{(2)}),
f_{\text{high}}(x_{\text{ring}}^{(3)})
\big].
\end{equation}

This step explicitly summarizes residual evidence across frequency ranges, preserving band-specific priors while reducing dimensionality before encoding.

\subsection{Autoencoder with Template Reconstruction}
\label{subsec:ae}
The refined vector $x_{\text{ref}}$ is encoded using an autoencoder with a one-dimensional latent representation. The encoder applies a small non-linear mapping to obtain the latent code $\hat{z}$, while the decoder maps $\hat{z}$ back to a ring-level reconstruction $\hat{x}_{\text{ring}} \in \mathbb{R}^{R}$. To recover the full ring matrix, we introduce a learnable template matrix $\mathbf{T} \in \mathbb{R}^{R \times K_{\max}}$. The reconstructed matrix is obtained through a factorized formulation:
\begin{equation}
\hat{\mathbf{X}} = \hat{x}_{\text{ring}} \odot \mathbf{T},
\end{equation}
where $\hat{x}_{\text{ring}}$ is broadcast along the frequency dimension. This structure constrains the reconstruction to respect the ring-wise organization while allowing frequency-specific patterns to be learned through the matrix $\mathbf{T}$.

\subsection{Training Objective}
\label{subsec:loss}
The network is trained by minimizing two reconstruction losses. The first enforces accurate reconstruction of the ring-level descriptor, while the second term encourages consistency between the reconstructed and original matrices and is applied only to valid coefficients.
\begin{equation}
\mathcal{L}=
\underbrace{\left\|\hat{\mathbf{x}}_{\text{ring}}-\mathbf{x}_{\text{ring}}\right\|_2^2}_{\substack{\text{ring-level}\\\text{reconstruction}\\\text{loss}}}
+
\underbrace{\frac{\left\|(\hat{\mathbf{X}}-\mathbf{X})\odot\mathbf{M}\right\|_2^2}{\sum \mathbf{M}}}_{\substack{\text{masked matrix}\\\text{reconstruction}\\\text{loss}}}.
\end{equation}

Optimizing these terms of the loss jointly is essential since the ring-level loss stabilizes the low-dimensional representation, while the full reconstruction loss ensures that gradients propagate through the learnable projection and template components, preventing degenerate solutions.

\section{Experimental Setup}

\begin{table*}[t]
\centering
\caption{Comparison with state-of-the-art one-class and self-supervised S-MAD methods trained on SMDD. We use metrics such as EER (\%), and BPCER@APCER (\%), reported at APCER $=5\%$ and $10\%$.}
\label{tab:placeholder_results}
\resizebox{\linewidth}{!}{%
\begin{tabular}{ll|ccc|ccc|ccc|ccc}
\hline
\multicolumn{2}{c|}{\multirow{3}{*}{\textbf{Test data}}} &
\multicolumn{3}{c|}{\textbf{SelfMAD++~\cite{selfmad++}}} &
\multicolumn{3}{c|}{\textbf{SPL-MAD~\cite{spl-mad}}} &
\multicolumn{3}{c|}{\textbf{MAD-DDPM~\cite{mad_ddpm}}} &
\multicolumn{3}{c}{\textbf{SRL-MAD (Ours)}} \\
\cline{3-14}
\multicolumn{2}{c|}{} &
\textbf{EER} & \multicolumn{2}{c|}{\textbf{BPCER@APCER (\%)}} &
\textbf{EER} & \multicolumn{2}{c|}{\textbf{BPCER@APCER (\%)}} &
\textbf{EER} & \multicolumn{2}{c|}{\textbf{BPCER@APCER (\%)}} &
\textbf{EER} & \multicolumn{2}{c}{\textbf{BPCER@APCER (\%)}} \\
\multicolumn{2}{c|}{} &
(\%) & 5.00 & 10.00 &
(\%) & 5.00 & 10.00 &
(\%) & 5.00 & 10.00 &
(\%) & 5.00 & 10.00 \\
\hline
\multirow{3}{*}{\textbf{FERET-Morph~\cite{frll_morph}}} &
FaceMorpher
& 0.43 & 0.38 & 0.19
& 20.42 & 65.39 & 40.85
& 27.98 & 99.24 & 95.27
& 16.47 & 37.89 & 25.62 \\
&
OpenCV
& 0.43 & 0.19 & 0.19
& 25.71 & 77.02 & 57.45
& 31.38 & 99.24 & 95.27
& 5.51 & 8.47 & 1.57 \\
&
StyleGAN2
& 1.49 & 2.08 & 0.19
& 25.33 & 82.34 & 62.06
& 32.14 & 99.24 & 95.27
& 1.92 & 0.60 & 0.45 \\
\hline

\multirow{5}{*}{\textbf{FRLL-Morph~\cite{frll_morph}}} &
AMSL
& 0.99 & 0.05 & 0.05
& 3.26 & 36.14 & 0.50
& 27.13 & 98.99 & 94.94
& 0.00 & 0.49 & 0.49 \\
&
FaceMorpher
& 0.00 & 0.26 & 0.26
& 1.03 & 0.99 & 0.99
& 10.40 & 99.14 & 95.19
& 0.12 & 0.49 & 0.49 \\
&
OpenCV
& 0.00 & 0.00 & 0.00
& 1.88 & 5.45 & 0.50
& 13.76 & 99.02 & 95.17
& 5.78 & 6.86 & 3.43 \\
&
StyleGAN2
& 1.48 & 2.13 & 0.25
& 14.65 & 64.85 & 32.18
& 14.32 & 99.10 & 95.17
& 13.01 & 25.98 & 15.20 \\
&
Webmorph
& 3.45 & 1.64 & 3.94
& 6.39 & 30.69 & 11.39
& 30.30 & 99.10 & 95.09
& 0.00 & 0.49 & 0.49 \\
\hline

\multicolumn{2}{c|}{\textbf{MorDIFF~\cite{mordiff}}}
& 0.00 & 0.00 & 0.00
& 9.78 & 54.95 & 23.27
& 2.80 & 99.20 & 96.40
& 0.00 & 0.00 & 0.00 \\
\hline

\multicolumn{2}{c|}{\textbf{Average}}
& \textbf{1.03} & \textbf{1.20} & \textbf{0.62}
& 14.27 & 46.20 & 28.91
& 23.58 & 99.12 & 95.60
& \underline{4.76} & \underline{9.03} & \underline{5.30} \\
\hline
\end{tabular}%
}
\end{table*}

\begin{table*}[t]
\centering
\caption{Comparison with state-of-the-art discriminative S-MAD methods trained on SMDD. We use metrics such as EER (\%), and BPCER@APCER (\%), reported at fixed APCER $=5\%$ and APCER $=10\%$.}
\label{tab:placeholder_results_discriminative}
\resizebox{\linewidth}{!}{%
\begin{tabular}{ll|ccc|ccc|ccc|ccc}
\hline
\multicolumn{2}{c|}{\multirow{3}{*}{\textbf{Test data}}} &
\multicolumn{3}{c|}{\textbf{MixFaceNet~\cite{smdd_inception_mixfacenet}}} &
\multicolumn{3}{c|}{\textbf{PW-MAD~\cite{pwmad}}} &
\multicolumn{3}{c|}{\textbf{MADation~\cite{caldeira2025madation}}} &
\multicolumn{3}{c}{\textbf{SRL-MAD (Ours)}} \\
\cline{3-14}
\multicolumn{2}{c|}{} &
\textbf{EER} & \multicolumn{2}{c|}{\textbf{BPCER@APCER (\%)}} &
\textbf{EER} & \multicolumn{2}{c|}{\textbf{BPCER@APCER (\%)}} &
\textbf{EER} & \multicolumn{2}{c|}{\textbf{BPCER@APCER (\%)}} &
\textbf{EER} & \multicolumn{2}{c}{\textbf{BPCER@APCER (\%)}} \\
\multicolumn{2}{c|}{} &
(\%) & 5.00 & 10.00 &
(\%) & 5.00 & 10.00 &
(\%) & 5.00 & 10.00 &
(\%) & 5.00 & 10.00 \\
\hline

\multirow{3}{*}{\textbf{FERET-Morph~\cite{frll_morph}}} &
FaceMorpher
& 19.99 & 78.26 & 53.69
& 8.36 & 89.79 & 42.34
& 7.08 & 15.69 & 9.26
& 16.47 & 37.89 & 25.62 \\
&
OpenCV
& 20.62 & 85.26 & 53.88
& 16.09 & 97.92 & 74.48
& 15.00 & 29.36 & 21.40
& 5.51 & 8.47 & 1.57 \\
&
StyleGAN2
& 36.64 & 96.79 & 87.33
& 20.91 & 98.30 & 81.47
& 21.66 & 46.31 & 36.86
& 1.92 & 0.60 & 0.45 \\
\hline

\multirow{5}{*}{\textbf{FRLL-Morph~\cite{frll_morph}}} &
AMSL
& 31.03 & 75.72 & 65.56
& 4.43 & 26.53 & 4.64
& 27.94 & 77.89 & 61.43
& 0.00 & 0.49 & 0.49 \\
&
FaceMorpher
& 8.37 & 23.63 & 11.51
& 1.97 & 3.18 & 1.72
& 1.47 & 1.31 & 0.74
& 0.12 & 0.49 & 0.49 \\
&
OpenCV
& 9.85 & 65.11 & 15.89
& 2.46 & 4.10 & 1.80
& 1.47 & 2.38 & 1.39
& 5.78 & 6.86 & 3.43 \\
&
StyleGAN2
& 38.92 & 94.44 & 84.94
& 17.24 & 49.51 & 34.62
& 19.61 & 47.55 & 32.57
& 13.01 & 25.98 & 15.20 \\
&
Webmorph
& 31.03 & 96.72 & 78.38
& 9.85 & 23.59 & 12.29
& 24.51 & 56.76 & 42.67
& 0.00 & 0.49 & 0.49 \\
\hline

\multicolumn{2}{c|}{\textbf{MorDIFF~\cite{mordiff}}}
& 5.88 & 8.80 & 6.00
& 11.27 & 43.40 & 18.40
& 24.02 & 50.20 & 38.00
& 0.00 & 0.00 & 0.00 \\
\hline

\multicolumn{2}{c|}{\textbf{Average}}
& 22.48 & 69.41 & 50.80
& \underline{10.29} & 48.48 & 30.20
& 15.86 & \underline{36.38} & \underline{27.15}
& \textbf{4.76} & \textbf{9.03} & \textbf{5.30} \\
\hline
\end{tabular}%
}
\end{table*}
\vspace{0.5em}
\noindent\textbf{Datasets.}
Although our approach could be adapted to arbitrary bona fide face datasets since our approach is one-class unsupervised, we strictly follow the protocols used in some recent MAD papers~\cite{selfmad, caldeira2025madation, synmad22} and train our method on SMDD~\cite{smdd_inception_mixfacenet} to ensure comparability with prior work. No morph samples are used during training, reflecting realistic deployment scenarios where attack samples may not be available during system development. Nevertheless, training on larger bona fide datasets (e.g., CASIA-WebFace) is an interesting direction to assess potential gains in generalization. To assess the generalization capability of the proposed method, as test data, we rely on three morphing datasets, FRLL-Morph, FERET-Morph~\cite{frll_morph}, and MorDIFF~\cite{mordiff}. SMDD is a synthetic StyleGAN2-based development dataset with OpenCV morphs, whereas FRLL-Morph is an FRLL-based benchmark containing landmark and GAN-based morphs being identity-disjoint from SMDD. MorDIFF~\cite{mordiff} provides additional FRLL-based diffusion morphs, enabling evaluation under stronger and more realistic attacks.

\vspace{0.5em}
\noindent\textbf{Evaluation metrics.}
All experiments follow the ISO/IEC~30107-3~\cite{ISO} standard for biometric presentation attack detection. Based on the anomaly scores, we report the Bona Fide Presentation Classification Error Rate (BPCER) at fixed Attack Presentation Classification Error Rate (APCER) thresholds of 5\% and 10\%. Additionally, the overall detection accuracy is reported as the Equal Error Rate (EER), where APCER equals BPCER.

\vspace{0.5em}
\noindent\textbf{Implementation Details.} We conduct the experiments on NVIDIA GeForce RTX 5090 containing 32GB of VRAM, and the framework utilized was Pytorch. To ensure consistency across evaluations, Fourier features were extracted using images of uniform size 500$\times$500, as previous works~\cite{pca_morph, selfmad++, paulo2026fdmadfrequencydomainresidualanalysis} suggest that higher-resolution images typically preserve more morphing artifacts, specifically in the Fourier domain. We optimize the model of our method with Adam with a learning rate of $5\times10^{-5}$ using a batch size of 128 for training.

\section{Results}
We decided to benchmark our method against both one-class and supervised MAD models. For consistency with recent MAD benchmarking studies and to ensure fair comparison across methods, we report baseline results following the evaluation protocol adopted in~\cite{selfmad++}. It is worth mentioning that unsupervised one-class methods operate under an open-set assumption, where only bona fide samples are available during training. This setting is well aligned with real deployments, where the space of possible morph generation pipelines is effectively unbounded. The main difficulty in such open-set scenarios is therefore the lack of knowledge of the attack types during training. In contrast, supervised approaches rely on explicit supervision on attack samples but are often biased toward the morph type present in training and degrade under cross-morph or dataset shifts. Despite this difference, our method achieves competitive results with one-class approaches and, in several comparisons against supervised approaches, achieves superior performance, confirming its potential for open-set morphing attack detection.

\vspace{0.5em}
\noindent\textbf{Comparison to One-Class and Self-Supervised MAD Models.} We compare our approach with SPL-MAD~\cite{spl-mad} and MAD-DDPM~\cite{mad_ddpm}, which represent recent baselines in one-class S-MAD, as well as SelfMAD++~\cite{selfmad++}, which is a self-supervised S-MAD method. While these methods differ in their modeling strategies and input representations, all approaches are trained under the same one-class constraint on SMDD and evaluated on the same target benchmarks using standard metrics. Results presented in Table~\ref{tab:placeholder_results} show that our method outperforms SPL-MAD~\cite{spl-mad} and MAD-DDPM~\cite{mad_ddpm} across all datasets, achieving a lower average EER of 4.76\% compared to \ 14.27\% and 23.58\%, respectively. In particular, MAD-DDPM~\cite{mad_ddpm} exhibits near-saturated BPCER values across most test sets, which indicates limited performance when forced to operate at a low APCER. Regarding SelfMAD++~\cite{selfmad++}, it achieves low error rates, which can be attributed to its use of explicit self-supervised artifact generation, which provides access to attack-related cues during training. Our method achieves its strongest gains on OpenCV and StyleGAN2 attacks from FERET-Morph~\cite{frll_morph}, obtaining EERs of 5.51\% and 1.92\%, respectively; however, performance degrades in more challenging scenarios, such as FRLL StyleGAN2. Nevertheless, our method remains competitive and consistently improves upon SPL-MAD~\cite{spl-mad} and MAD-DDPM~\cite{mad_ddpm}. Overall, these results highlight the effectiveness of modeling frequency-domain residuals as a domain-agnostic cue for morph detection.

\vspace{0.5em}
\noindent\textbf{Comparison to Supervised MAD Models.} We further compare our method with state-of-the-art supervised approaches, including MixFaceNet~\cite{smdd_inception_mixfacenet}, PW-MAD~\cite{pwmad}, and MADation~\cite{caldeira2025madation}, as reported in Table~\ref{tab:placeholder_results_discriminative}. These methods rely on supervised training with morph samples, which can improve performance when the training and testing morph distributions are similar. However, such models may also overfit to the morph generation process observed during training and degrade under cross-dataset evaluation. Despite this, our method achieves the best overall performance, with an average EER of 4.76\% compared to 10.29\% for PW-MAD~\cite{pwmad} and 15.86\% for MADation~\cite{caldeira2025madation}, and dramatically lower BPCER values at both APCER operating points (9.03\% and 5.30\%, respectively). On AMSL and Webmorph attacks from FRLL-Morph~\cite{frll_morph}, our method achieves EERs of 0.00\%, while supervised approaches exhibit significantly higher EERs. This observation supports the advantage of one-class models, which avoid overfitting to specific morph attacks. Overall, the proposed approach advances the state of the art among one-class MAD methods and surpasses several supervised models, despite relying solely on bona fide training data.

\section{Conclusion}
This paper describes a novel one-class method for detecting face morphing attacks. Our method uses residual Fourier features and learns a compact latent representation for anomaly scoring. Through extensive benchmarking across multiple datasets and comparisons with other one-class and supervised approaches, we show that our method achieves competitive performance and even surpasses common supervised approaches, showing the potential of using residual Fourier features as a proxy for morphing artifacts. Future work includes training on larger datasets, such as CASIA-WebFace, to capture more bona fide patterns and to further improve open-set robustness.

\section*{Acknowledgments}
This work was funded by the Portuguese Foundation for Science and Technology (FCT) under the PhD grant 2025.03420.BD. This work is supported by NOVA LINCS (UID/04516/2025) with the financial support of FCT.IP (DOI: \url{https://doi.org/10.54499/UID/04516/2025}). It was also supported, when eligible, by national funds through FCT under project UID/50008/2025 – Instituto de Telecomunicações (DOI: \url{https://doi.org/10.54499/UID/50008/2025}).

\bibliography{references}
\bibliographystyle{ieeetr}
\end{document}